\documentclass[10pt,twocolumn,letterpaper]{article}

\usepackage{cvpr}              

\usepackage{times}
\usepackage{epsfig}
\usepackage{graphicx}
\usepackage{amsmath}
\usepackage{amssymb}

\usepackage{algorithm}
\usepackage{algorithmic}
\usepackage{mathrsfs}

\usepackage{graphicx}
\usepackage{times}
\usepackage{epsfig}
\usepackage{graphicx}
\usepackage{amsmath}
\usepackage{amssymb}
\usepackage{caption}
\usepackage{subcaption}
\usepackage{tabularx}
\usepackage{rotating}
\usepackage{verbatim}
\usepackage{xcolor,colortbl}
\usepackage{etoolbox}
\usepackage[normalem]{ulem}
\usepackage{booktabs}
\usepackage{multirow}
\usepackage{lipsum}
\usepackage{stmaryrd}
\usepackage{stackengine}
\usepackage{makecell}
\usepackage{pifont}
\usepackage{cancel}
\usepackage{adjustbox}
\usepackage{dblfloatfix}
\usepackage{bbding}
\usepackage{pifont}
\usepackage{wasysym}
\usepackage{amssymb}
\usepackage{bm}
\definecolor{car}{rgb}{0.39215686, 0.58823529, 0.96078431}
\definecolor{bicycle}{rgb}{0.39215686, 0.90196078, 0.96078431}
\definecolor{motorcycle}{rgb}{0.11764706, 0.23529412, 0.58823529}
\definecolor{truck}{rgb}{0.31372549, 0.11764706, 0.70588235}
\definecolor{other-vehicle}{rgb}{0.39215686, 0.31372549, 0.98039216}
\definecolor{person}{rgb}{1.        , 0.11764706, 0.11764706}
\definecolor{bicyclist}{rgb}{1.        , 0.15686275, 0.78431373}
\definecolor{motorcyclist}{rgb}{0.58823529, 0.11764706, 0.35294118}
\definecolor{road}{rgb}{1.        , 0.        , 1.        }
\definecolor{parking}{rgb}{1.        , 0.58823529, 1.        }
\definecolor{sidewalk}{rgb}{0.29411765, 0.        , 0.29411765}
\definecolor{other-ground}{rgb}{0.68627451, 0.        , 0.29411765}
\definecolor{building}{rgb}{1.        , 0.78431373, 0.        }
\definecolor{fence}{rgb}{1.        , 0.47058824, 0.19607843}
\definecolor{vegetation}{rgb}{0.        , 0.68627451, 0.        }
\definecolor{trunk}{rgb}{0.52941176, 0.23529412, 0.        }
\definecolor{terrain}{rgb}{0.58823529, 0.94117647, 0.31372549}
\definecolor{pole}{rgb}{1.        , 0.94117647, 0.58823529}
\definecolor{traffic-sign}{rgb}{1.        , 0.        , 0.    }

\makeatletter
\newcommand{\car@semkitfreq}{3.92}
\newcommand{\bicycle@semkitfreq}{0.03}
\newcommand{\motorcycle@semkitfreq}{0.03}
\newcommand{\truck@semkitfreq}{0.16}
\newcommand{\othervehicle@semkitfreq}{0.20}
\newcommand{\person@semkitfreq}{0.07}
\newcommand{\bicyclist@semkitfreq}{0.07}
\newcommand{\motorcyclist@semkitfreq}{0.05}
\newcommand{\road@semkitfreq}{15.30}  %
\newcommand{\parking@semkitfreq}{1.12}
\newcommand{\sidewalk@semkitfreq}{11.13}  %
\newcommand{\otherground@semkitfreq}{0.56}
\newcommand{\building@semkitfreq}{14.1}  %
\newcommand{\fence@semkitfreq}{3.90}
\newcommand{\vegetation@semkitfreq}{39.3}  %
\newcommand{\trunk@semkitfreq}{0.51}
\newcommand{\terrain@semkitfreq}{9.17} %
\newcommand{\pole@semkitfreq}{0.29}
\newcommand{\trafficsign@semkitfreq}{0.08}
\newcommand{\semkitfreq}[1]{{\csname #1@semkitfreq\endcsname}}

\usepackage{enumitem} 
\setlist[itemize]{nosep, topsep=0pt, partopsep=0pt, leftmargin=*}

\definecolor{cvprblue}{rgb}{0.21,0.49,0.74}
\usepackage[pagebackref,breaklinks,colorlinks,citecolor=cvprblue]{hyperref}

\definecolor{barrier}{RGB}{112,128,144}
\definecolor{bicycle}{RGB}{220,20,60}
\definecolor{bus}{RGB}{255, 127, 80}
\definecolor{car}{RGB}{255, 158, 0}
\definecolor{const. veh.}{RGB}{233, 150, 70}
\definecolor{motorcycle}{RGB}{255,61,99}
\definecolor{pedestrian}{RGB}{0,0,230}
\definecolor{traffic cone}{RGB}{47,79,79}
\definecolor{trailer}{RGB}{255,140,0}
\definecolor{truck}{RGB}{255,99,71}
\definecolor{drive. suf.}{RGB}{0,207,191}
\definecolor{other flat}{RGB}{175,0,75}
\definecolor{sidewalk}{RGB}{75,0,75}
\definecolor{terrain}{RGB}{112,180,60}
\definecolor{manmade}{RGB}{222,184,135}
\definecolor{vegetation}{RGB}{0,175,0}

\def\paperID{5516} 
\def\confName{CVPR}
\def\confYear{2024}

\title{SparseOcc: Rethinking Sparse Latent Representation for \\
Vision-Based Semantic Occupancy Prediction}
\author{
    Pin Tang$^{1}$ \quad Zhongdao Wang$^{2}$ \quad Guoqing Wang$^{1}$ \quad Jilai Zheng$^{1}$ \\
    \quad Xiangxuan Ren$^{1}$ \quad Bailan Feng$^{2}$ \quad Chao Ma$^{1}$\thanks{~Corresponding author.}\\ 
    ${}^{1}$ MoE Key Lab of Artificial Intelligence, AI Institute, Shanghai Jiao Tong University \\
    ${}^{2}$ Huawei Noah's Ark Lab\\
    {\tt\small \{pin.tang,guoqing.wang,zhengjilai,bunny\_renxiangxuan,chaoma\}@sjtu.edu.cn} \\
    {\tt\small \{wangzhongdao,fengbailan\}@huawei.com}\\
    {\small Project page: \url{https://pintang1999.github.io/sparseocc.html}}
    }

\begin{document}
\maketitle
\begin{abstract}
Vision-based perception for autonomous driving requires an explicit modeling of a 3D space, where 2D latent representations are mapped and subsequent 3D operators are applied. However, operating on dense latent spaces introduces a cubic time and space complexity, which limits scalability in terms of perception range or spatial resolution. Existing approaches compress the dense representation using projections like Bird's Eye View (BEV) or Tri-Perspective View (TPV). Although efficient, these projections result in information loss, especially for tasks like semantic occupancy prediction.
To address this, we propose SparseOcc, an efficient occupancy network inspired by sparse point cloud processing. It utilizes a lossless sparse latent representation with three key innovations. Firstly, a 3D sparse diffuser performs latent completion using spatially decomposed 3D sparse convolutional kernels. Secondly, a feature pyramid and sparse interpolation enhance scales with information from others. Finally, the transformer head is redesigned as a sparse variant.
SparseOcc achieves a remarkable \textbf{74.9\%} reduction on FLOPs over the dense baseline. Interestingly, it also improves accuracy, from 12.8\% to 14.1\% mIOU, which in part can be attributed to the sparse representation's ability to avoid hallucinations on empty voxels.

\end{abstract}    
\section{Introduction}
\label{sec:intro}
\begin{figure}
    \centering
    \includegraphics[scale=.5]{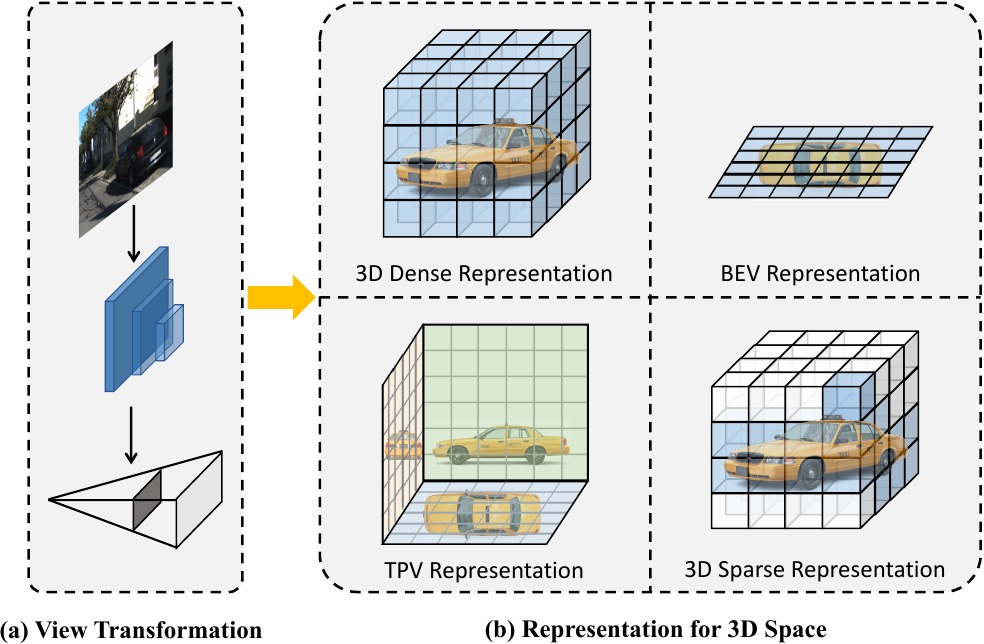}
    \caption{{(a)}  Vision-based perception methods for autonomous driving typically first extract image features by a 2D latent encoder and then map them to  3D  using view transformation. 
    {(b)} For the 3D latent space, existing methods mostly employ the dense,
    BEV,
    or TPV
    representation, while we rethink the possibility of using sparse representation to achieve superior efficiency and accuracy.}
    \label{fig:sparse_vs_dense}
    \vspace{-10pt}
\end{figure}
Accurate perception of the surrounding environment is crucial for autonomous driving systems~\cite{jia2023driveadapter,uniad}. In recent years, vision-based 3D perception algorithms have gained significant attention and advancement due to their cost-effectiveness. The typical workflow involves employing a 2D encoder to extract latent representations from images. A view transformation method, such as lift-splat-shoot (LSS)~\cite{lss}, is then applied to lift the perspective 2D latent features to a 3D voxel space, utilizing predicted depth information.
This 3D scene representation serves as the foundation for deriving geometry and semantic information to describe the driving environment, supporting various 3D perception tasks including object detection~\cite{bevdepth, pointpillar, zhang2024alleviating, zhou2023unidistill}, semantic segmentation~\cite{uniad, rclane, tang2023prototransfer}, and semantic occupancy prediction~\cite{surroundocc, tpvformer, OpenOccupancy}. In this study, we specifically focus on the challenging task of semantic occupancy prediction~\cite{OpenOccupancy, surroundocc, tong2023scene, tian2023occ3d}, which entails predicting both static and dynamic elements within the scene.

Several alternatives are available for representing 3D spatial latent information. The dense representation~\cite{OpenOccupancy} is the most straightforward approach, storing features in continuous memory and enabling direct application of dense 3D operations like convolutions. However, this representation is redundant and inefficient, as approximately 67\% of the considered 3D space is empty\footnote{We conduct statics using the ground-truth occupancy labels of the first 10 sequences in the SemanticKITTI dataset.}.
Bird Eye's View (BEV)~\cite{bevdet,bevfusion,BEVFormer} has gained popularity as a recent prevalent representation. It involves projecting the 3D space onto a BEV plane, significantly reducing computational costs by leveraging efficient 2D building blocks in a BEV encoder. However, this projection introduces the loss of geometry information, thereby limiting the fine-grained capacity of the BEV representation to comprehend the 3D scene structure.
A Tri-Perspective View (TPV)~\cite{tpvformer} representation is proposed to mitigate the information loss, but it still suffers from degraded perception accuracy.

In this paper, we seek a latent representation that encodes the 3D scene structure in a lossless manner while minimizing computational costs.
Drawing inspiration from the similarities between sparse 3D vision features and point clouds, we rethink the feasibility of employing a pure sparse representation for the 3D latent space, a common practice in point cloud processing~\cite{spconv}. 
Specifically, we utilize the coordinate (COO) format to store sparse tensors and introduce a series of sparse building blocks tailored for the sparse representation. Our proposed method, SparseOcc, is an occupancy network where all 3D layers operate on sparse tensors.
The key designs of SparseOcc are as follows:
\begin{itemize}
    \item \textbf{Sparse Latent Diffuser:} This component enables the propagation of non-empty features to adjacent empty regions, facilitating scene completion. To ensure efficiency, a 3D diffusion kernel is spatially decomposed into a combination of three orthogonal convolutional kernels.
    \item \textbf{Sparse Feature Pyramid:} We build a feature pyramid that incorporates sparse interpolation operations to enhance scales with information from other scales. This pyramid design expands the reception fields, reducing the need for excessive diffusers within each scale, which helps preserve sparsity. 
    \item \textbf{Sparse Transformer Head:} The final component of SparseOcc is a 3D sparse transformer head responsible for generating semantic occupancy predictions. By segmenting only occupied voxels rather than the entire 3D volume, we achieve a remarkable reduction in computational costs.
\end{itemize}

Based on the sparse latent representation, SparseOcc achieves a significant reduction in computation overhead. In the nuScenes-Occupancy benchmark~\cite{OpenOccupancy},  it reduces the FLOPs of existing approaches using dense or TPV representations by \textbf{59.8\% to 74.9\%}, and memory usage by \textbf{31.6\% to 40.9\%}. 
Remarkably, the semantic occupancy accuracy not only remains intact but improves, surpassing the state-of-the-art C-CONet. Specifically, the semantic occupancy accuracy increases from 12.8\% to 14.1\% mIoU. This improvement highlights the superiority of the sparse representation over the dense one in terms of accuracy, as it naturally avoids hallucinations on empty voxels.
Considering the effectiveness and efficiency demonstrated by SparseOcc, we propose that it can serve as a new baseline for occupancy networks.
\section{Related Work}
\subsection{3D Scene Representation}
\noindent\textbf{3D Dense Representation.}
Representing the surrounding environment with spatial latent information is an indispensable procedure for autonomous driving perception algorithms. 
A straightforward solution is to split the scene into voxels and describe the scene with 3D volume representation where 3D operators are applied~\cite{sscnet,OpenOccupancy, surroundocc, occformer,monoscene}. For example, OpenOccupancy~\cite{OpenOccupancy} first uses a 2D encoder to extract image features and further uses LSS to lift the features to 3D space. Then, ResNet3D~\cite{resnet} and FPN3D~\cite{fpn} are used to diffuse non-empty features to adjacent empty areas. SurroudOcc~\cite{surroundocc} uses successive deformable cross attention layers~\cite{deformable_DETR} to transform the multi-scale image features to multi-scale 3D dense volume. However, these 3D operators are often in cubic time and spatial time complexity, which is not affordable in practice.

\noindent\textbf{BEV Representation.}
The past several years have witnessed the prosperous development of BEV representation in tasks such as 3D object detection~\cite{bevfusion, BEVFormer, metabev, bevdepth}, BEV semantic segmentation~\cite{bevsegformer, CVT, pon}, and instance prediction~\cite{hu2021fiery}. They either forward project the image features to 3D spaces with estimated depth, then compress the 3D feature volume to BEV map, or update the BEV queries via backward deformable attention. Despite efficiency, the geometry lossy projection results in the relatively coarse representation of the 3D scene, hindering its generalization to the fine-grained semantic occupancy prediction task.

\noindent\textbf{TPV Representation.}
To make a trade-off between efficacy and efficiency, TPV representation is proposed~\cite{tpvformer, zuo2023pointocc}. It first constructs three cross-planes perpendicular to each other to represent the 3D scene. Then, sets of queries are initialized on these planes to aggregate features from images and exchange features across views via the attention mechanism. Then, they efficiently reconstruct each voxel in the 3D space by summing its projected features on the three planes for downstream tasks. Although the loss of geometry is mitigated, it is still hard to capture the complex driving scene, leading to degraded performance.

In this paper, we rethink the 3D scene representation from a sparse view and seek an effective but also efficient method for semantic occupancy prediction. 

\subsection{Scene Completion in Occupancy Prediction}
Both LiDAR points and lifted 3D representation only cover the initial intersection surface between the ray from the sensor and objects. Hence, a few methods have been proposed to diffuse non-empty features to surrounding empty regions~\cite{SCPNet, voxformer}. 
These methods essentially are the same as the 3D dense representation based ones since they also stack 3D dense operators to diffuse features. 
In contrast, we explore the potential of using a pure sparse representation and propose a sparse latent diffuser for scene completion. This approach rethinks the conventional method of feature diffusion in 3D space, offering a more efficient way to fill in the gaps in the scene representation.
\section{Proposed Approach}
\begin{figure*}
    \centering
    \includegraphics[width=\linewidth]{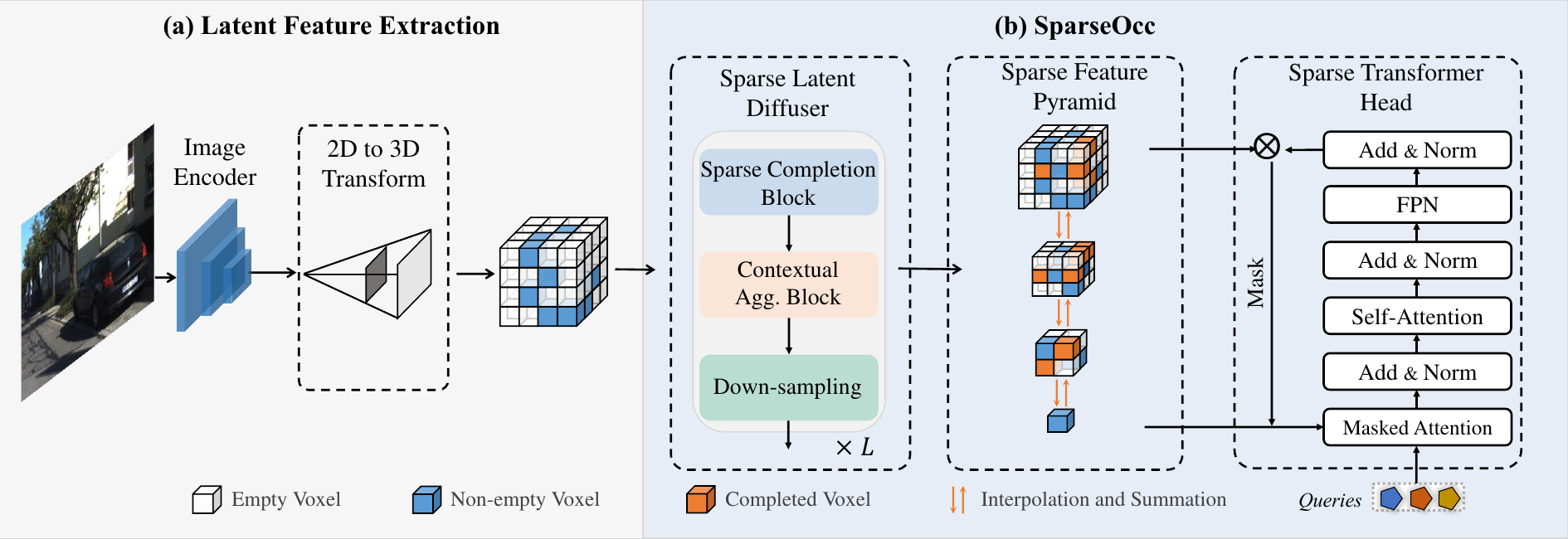}
    \caption{
    \textbf{Overview of the proposed approach.} (a) Images captured by monocular or surrounding cameras are first passed to a 2D encoder, yielding 2D latent features. Then the latent features are mapped to 3D using the predicted depth map following the LSS~\cite{lss}. (b) SparseOcc adopts a sparse representation for the latent space. Upon this representation, we introduce three key building blocks: a latent diffuser that performs completion, a feature pyramid that enhances receptive filed, and a transformer head that predicts semantic occupancy. 
    }
    \label{fig:sparseocc}
\end{figure*}
\subsection{Preliminary: Sparse Representation}
\label{sec:view_transform}
Given a monocular image or a set of images  $\mathbf{I}=\{I_i\}_{i=1}^{N_\text{view}}$ captured by surrounding cameras, the goal of vision-based 3D semantic occupancy prediction is to predict a semantic label for every voxel in a predefined 3D volume $\{0,1,2,...,C\}^{H\times W \times D}$, where $H,W,D$ denotes the spatial resolution, the class label 0 indicates empty voxel, $C$ is the number of semantic classes for non-empty voxels.

In the initial stage of our model, we follow the Lift-Splat-Shoot (LSS)~\cite{lss} framework.
The images are first passed through an image encoder, such as ResNet~\cite{resnet} augmented with FPN~\cite{fpn}, This encoder generates latent features on the 2D perspective plane. Subsequently, the 2D latent features are lifted to the 3D space using predicted depth maps, resulting in dense cubic features denoted as $\mathbf{V}\in \mathbb{R}^{H\times W \times D \times C}$. Notably, we observe that approximately 80\% of the voxels in $\mathbf{V}$ are empty. This sparsity arises due to the nature of LSS, where 2D features are cast through ray casting and become sparser in distant regions. 

We then convert the dense feature $\mathbf{V}$ to a sparse representation by gathering non-empty voxels. The sparse tensor is stored in a commonly used coordinate (COO) format:
$$
\mathbb{V} = \{ (\mathbf{p}_i=[x_i, y_i, z_i] \in \mathbb{R}^3, \mathbf{f}_i \in \mathbb{R}^C )  \vert i=1,2,...N  \}.
$$

In the above equation, $N$ represents the number of non-empty voxels, while $\mathbf{p}_i$ and $\mathbf{f}_i$ denote the coordinates and features of the $i$-th voxel, respectively. All subsequent operations are performed on this sparse representation, eliminating redundant computations on empty voxels. The following sections elaborate on the proposed sparse building blocks, namely the sparse latent diffuser, feature pyramid, and transformer head. An overview is shown in Fig.~\ref{fig:sparseocc}.

\subsection{Sparse Latent Diffuser}
\label{sec:sparse_latent_diffuser}
The sparse representation $\mathbb{V}$ is derived through a ray casting manner, resulting in a predominantly sparse depiction limited to the initial intersection face between a ray and an object. Consequently, the majority of observations are inherently incomplete. Contrarily, the objective of an occupancy network is to predict complete occupancy rather than solely the visible parts. Traditional approaches address this by incorporating 3D dense convolutions (such as 3D ResNet and 3D FPN) or attention layers (such as deformable self-attention) to diffuse non-empty features to adjacent empty regions, thereby completing the scene. In this work, we aim to design a sparse variant of the latent diffuser. However, a notable challenge arises as the objective of the diffuser appears to conflict with the sparse design: By stacking more completion blocks, the scene is better completed, but the spatial sparsity also decreases, hindering efficiency.
To strike a balance between scene completion and sparsity, we build our sparse latent diffuser with two key components: A sparse completion block, which executes only  \emph{necessary} latent diffusion; and a contextual aggregation block, which aggregates valid features \emph{without} engaging in completion.

\noindent\textbf{Sparse Completion Block.} 
We opt for the 3D sparse convolution implemented by~\cite{spconv} to build the sparse completion block. A sparse convolution performs the computation in a local window in which at least one non-empty voxel resides, allowing the diffusion of features from non-empty voxels to their neighbors. 
The range of diffusion can be expanded by stacking multiple layers of sparse convolution. 
To maintain the spatial sparsity, we only use one 3D convolution layer in a sparse completion block.

\noindent\textbf{Contextual Aggregation Block.}
After completion, we introduce the contextual aggregation block to effectively utilize geometry and semantic features from the local context. For constructing this block, we choose sparse submanifold convolution~\cite{spconv} over regular sparse convolution. Submanifold convolution ensures that an output location is active only if the corresponding input location is active, thereby maintaining sparsity even when stacking multiple layers.

\noindent\textbf{Kernel Decomposition.}
Foreground objects and background elements in driving scenes often exhibit specific shape distributions. For instance, roadways and sidewalks typically have a thin, flat shape located at the bottom of the 3D volume, making them amenable to completion through convolutions in the horizontal direction. Conversely, structures like buildings or car-like objects have a rectangular shape, necessitating feature diffusion in the vertical direction. To fully leverage these distinct shape distributions, we decompose a conventional $k\times k \times k$ kernel into orthogonal kernels~\cite{Cylinder3D}.
Specifically, for the sparse completion block, we replace the sparse convolution with three consecutive layers with $k\times k\times 1$, $k\times 1 \times k$, and $1 \times k \times k$ kernels, respectively.
For the contextual aggregation block, we follow Cylinder3D~\cite{Cylinder3D} and replace a $k\times k \times k$ submanifold convolution with two parallel but asymmetrical branches of decomposed layers. One branch consists of two consecutive layers with $1 \times k \times k$ ad  $k \times 1 \times k$ kernels, and the other branch with  $k \times 1 \times k$ ad  $1 \times k \times k$ kernels. Note that the complexity is reduced from $\mathcal{O}(k^3)$ to $\mathcal{O}(3k^2)$ or $\mathcal{O}(4k^2)$ after the decomposition. Though the actual cost is not reduced when we use a small kernel with $k=3$, the expressive capacity of decomposed kernels outperforms a single full kernel, thus we can still achieve efficiency improvements by stacking fewer layers.

\begin{figure}
    \centering
    \includegraphics[width=0.9\linewidth]{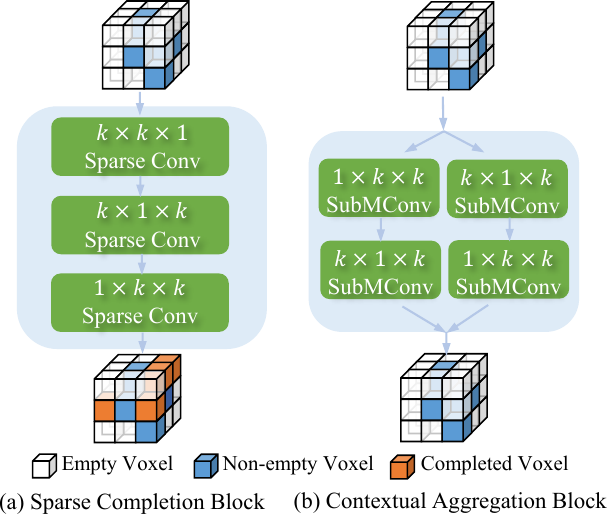}
    \caption{\textbf{Two building blocks of the sparse latent diffuser.} (a) The sparse completion block diffuses non-empty features to empty neighbors, and (b) The contextual aggregation block aggregates geometry and semantic features without engaging in completion.}
    \label{fig:scb}
    \vspace{-5pt}
\end{figure}

\subsection{Sparse Feature Pyramid}
A straightforward approach to complete the scene is to stack the proposed sparse diffuser multiple times, akin to SCPNet~\cite{SCPNet}. However, this necessitates a substantial number of sparse diffusers to ensure an adequately large receptive field, which is particularly important for recognizing large objects like ``truck'' or static elements such as ``road''. 
The computation cost is obviously expensive. 
To address this issue, we note that down-sample layers, implemented with sparse convolution with a stride greater than one, not only reduce the spatial resolution but also increase the relative sparsity in the new scale.
By building a multi-scale sparse feature pyramid with down-sample layers, we are readily to obtain a coarse-to-fine representation of the scene. This ensures that querying any spatial location can be addressed by at least one feature scale, simultaneously reducing computation costs.
Formally, we stack the sparse diffuser for $L$ times, each is followed by a down-sample layer, and the feature pyramid is collected as $\{\mathbb{V}_l\}_{l=1}^L$.
Additionally, the spatial size along the height dimension of the last two scales is too small, so we simply omit the $D$ dimension. For the completion blocks, the 3D convolution is replaced with a 2D version with $3\times 3$ kernel.
For the contextual aggregation block, we replace the asymmetrical branches with two parallel 2D submanifold convolutions with $5\times5$ kernels.

\noindent\textbf{Sparse Voxel Decoder.}
\label{sec:voxdecoder}
 Former methods~\cite{mask2former, occformer} uses multi-scale deformable attention transformer (MSDeformAttn)~\cite{deformabledetr} in the pixel/voxel decoder responsible for aggregating intra-scale and inter-scale features. Targeting for saving GPU memory and time, we simplify this process by using interpolation to fuse multi-scale sparse features output by the 3D sparse encoder. 
 Specifically, for a given scale $\mathbb{V}_l$ from the feature pyramid, we augment it by fusing all the other scales, 
\begin{equation}
    \hat{\mathbb{V}}_l = \sum_{j\ne l} W_j \cdot \mathrm{Interp}(\mathbb{V}_j, \mathbb{V}_l),2
\end{equation}
 where $W_j$ is a learned weight for the $j$-th scale, and $\mathrm{Interp}(\mathbb{X},\mathbb{Y})$ indicates linear interpolation from a sparse tensor $\mathbb{X}$ to another $\mathbb{Y}$. 

Leveraging this lightweight feature fusion approach, the feature pyramid is enriched with semantic information from different scales. Moreover, the high-resolution features benefit from additional completion provided by the low-resolution features, as the denser low-resolution features resist dilution by interpolation operators.

\subsection{Sparse Transformer Head}
\label{sec:transdecoder}
We frame the semantic occupancy prediction as a 3D segmentation problem and employ a transformer head, inspired by the design of Mask2Former~\cite{mask2former}. This head iteratively updates a set of learnable queries through masked attention and subsequently decodes these queries into 3D masks for all semantic classes. 
However, a challenge arises due to the need for dense binary masks to represent semantic classes. Additionally, as we perform predictions in a 3D space~\cite{occformer}, the computational and memory costs associated with applying successive transformer decoder layers to dense masks become impractical.

To overcome this challenge, we adapt the dense transformer head to a sparse variant. Specifically, we assert that accurate segmentation is crucial for occupied voxels, while non-occupied voxels need not be considered during this phase. In the ensuing discussion, we outline several steps to delineate the sparse transformer decoder.

\noindent\textbf{Preprocessing and Notations.}
Given the sparse feature pyramid $\hat{\mathbb{V}}_l$, we first employ a linear binary classifier for coarse segmentation.  
The binary classifier is trained to label a voxel as empty if the semantic ground truth is 0 and non-empty otherwise.
We only preserve voxels that are classified as non-empty, resulting in a filtered, sparser tenser $\hat{\mathbb{V}}_l = \{ (\mathbf{p}_i, \mathbf{f}_i) \vert i=1,...,N_l\}$, where we use $N_l$ to denote the number of remaining voxels for the $l$-th scale. 
Storing variant features for the empty voxels is unnecessary, and instead, we find it suffices to use only a single learnable token $\mathbf{p}_{\phi}$ to represent all the empty voxels.

\noindent\textbf{Query Decoding.} 
The former method~\cite{occformer} performs outer product between queries $Q \in \mathbb{R}^{N_q \times C}$ and 3D dense features $F \in \mathbb{R}^{C\times H\times W\times D}$ to decode the queries to 3D mask with shape $(N_q, H, W, D)$, which has a time complexity of $\mathcal{O}(N_q H W D C)$. 
To facilitate inference speed, we only perform outer product between $Q$ and $ \mathbf{p}_{\phi} \cup \{\mathbf{f}_i \vert (\mathbf{p}_i, \mathbf{f}_i) \in \hat{\mathbb{V}}_l \}$ to obtain a series of occupied masks $M^\mathrm{occ}\in\mathbb{R}^{N_q\times N_l}$ and a single mask $M^{\phi}\in\mathbb{R}^{N_q\times 1}$ that represents empty voxels. 
With the saved coordinates $\mathbf{p}$ of occupied voxels, we can easily reconstruct the dense 3D mask $M\in\mathbb{R}^{Q\times H\times W\times Z}$ from the predicted sparse tensor, using a scatter operation that does not break the gradient flow.
This way, the complexity is reduced to $\mathcal{O}(N_l N_q C+HWZ)$ for mask prediction and reconstruction.
Note that
\begin{align}\label{Eq:ema}
    & \mathcal{O}(N_l N_q C+HWZ)  \nonumber \\
    & = \mathcal{O}(HWZN_q C(\frac{N_l}{HWZ}+\frac{1}{N_qC}))  \\ 
    & < \mathcal{O}(H W Z N_q C), \nonumber
\end{align}
because ${N_l}/{HWZ}$ is rather smaller that $1$ in our sparse case.
Moreover, the $N_q$ queries are also input to a $C$-way linear classifier to classify the corresponding 3D mask into predefined $C$ semantic categories.

\noindent\textbf{Query Updating.}
Given the input $L$ layers of multi-scale sparse features, we iteratively alternate between query decoding and query updating in each transformer layer. 
With the predicted 3D masks $M_{l-1}$ in the $(l-1)$-th transformer layer, we update the queries via
\begin{equation}
    Q_l = \mathrm{softmax}\left[\mathcal{M}_{l-1}+W_q Q_{l-1}(W_k\hat{\mathbf{V}}_l)^T\right]W_v \hat{\mathbf{V}}_l+Q_{l-1},
\end{equation}
where $W_q, W_k, W_v$ are linear layers, $\hat{\mathbf{V}}_l$ is the dense version reconstrcuted from $\hat{\mathbb{V}}_l$, and the attention mask $\mathcal{M}_{l-1}$ at location $(x, y, z)$ is obtained by
\begin{align}
    &\mathcal{M}_{l-1}(x, y,z)=\left\{
\begin{array}{ll}
0  & \text{if } \sigma \left(M_{l-1}^{'}(x, y, z)\right) \geq 0.5\\
-\infty    & \text{otherwise}
\end{array} \right.
\end{align}
where $\sigma$ is the sigmoid function, $M_{l-1}^{'}=\mathrm{maxpooling}(M_{l-1})$ which resizes the 3D mask to the same resolution of $\hat{\mathbf{V}}_l$, similar to implementation in~\cite{occformer}.

\subsection{Objective Function}
Considering the sparse transformer head formulates semantic occupancy as a mask set prediction task~\cite{mask2former,occformer}, bipartite matching with Hungarian solver is used to assign binary mask labels and corresponding semantic class labels to the predicted masks. 
Based on the assignment, we calculate the mask loss $\mathcal{L}_\mathrm{mask}$ and classification loss $\mathcal{L}_\mathrm{cls}$. 
Additionally, $\mathcal{L}_\mathrm{depth}$ is calculated between the predicted depth map and ground-truth projected by point clouds, for supervision of the LSS component. Moreover, the coarse binary classification on non-empty voxels is also supervised by a segmentation loss $\mathcal{L}_\mathrm{seg}$.
Finally, the overall objective function is a simple summation of these loss terms
\begin{equation}
\mathcal{L} =  \mathcal{L}_\mathrm{mask}+\mathcal{L}_\mathrm{cls}+\mathcal{L}_\mathrm{depth}+\mathcal{L}_\mathrm{seg}.
\end{equation}
\section{Experiments}

\begin{table*}
        \scriptsize
	\setlength{\tabcolsep}{0.0035\linewidth}
	\newcommand{\classfreq}[1]{{~\tiny(\semkitfreq{#1}\%)}}  %
	\centering
   \resizebox{1\linewidth}{!}
   {
	\begin{tabular}{l|c| c c | c c c c c c c c c c c c c c c c | c | c | c}
 
		\hline
		Method
		& \makecell[c]{Input}
		& \makecell[c]{IoU}
            & \makecell[c]{mIoU}
		& \rotatebox{90}{\textcolor{barrier}{$\blacksquare$} barrier} 
		& \rotatebox{90}{\textcolor{bicycle}{$\blacksquare$} bicycle}
		& \rotatebox{90}{\textcolor{bus}{$\blacksquare$} bus} 
		& \rotatebox{90}{\textcolor{car}{$\blacksquare$} car} 
		& \rotatebox{90}{\textcolor{const. veh.}{$\blacksquare$} const. veh.} 
		& \rotatebox{90}{\textcolor{motorcycle}{$\blacksquare$} motorcycle} 
		& \rotatebox{90}{\textcolor{pedestrian}{$\blacksquare$} pedestrian} 
		& \rotatebox{90}{\textcolor{traffic cone}{$\blacksquare$} traffic cone} 
		& \rotatebox{90}{\textcolor{trailer}{$\blacksquare$} trailer} 
		& \rotatebox{90}{\textcolor{truck}{$\blacksquare$} truck} 
		& \rotatebox{90}{\textcolor{drive. suf.}{$\blacksquare$} drive. suf.} 
		& \rotatebox{90}{\textcolor{other flat}{$\blacksquare$} other flat} 
		& \rotatebox{90}{\textcolor{sidewalk}{$\blacksquare$} sidewalk} 
		& \rotatebox{90}{\textcolor{terrain}{$\blacksquare$} terrain} 
		& \rotatebox{90}{\textcolor{manmade}{$\blacksquare$} manmade} 
		& \rotatebox{90}{\textcolor{vegetation}{$\blacksquare$} vegetation}
            & \makecell[c]{FLOPs}
            & \makecell[c]{Memory}
            & \makecell[c]{3D/Overall \\Latency}
            \\
		\hline\hline
        MonoScene~\cite{monoscene} & C & 18.4 & 6.9 & 7.1  & 3.9  &  9.3 &  7.2 & 5.6  & 3.0  &  5.9& 4.4& 4.9 & 4.2 & 14.9 & 6.3  & 7.9 & 7.4  & \textbf{10.0} & 7.6 & - & - & -\\
  
        TPVFormer~\cite{tpvformer} &C & 15.3 &  7.8 & 9.3  & 4.1  &  11.3 &  10.1 & 5.2  & 4.3  & 5.9 & 5.3&  6.8& 6.5 & 13.6 & 9.0  & 8.3 & 8.0  & 9.2 & 8.2 & 1132G & 20G & 0.57/0.73s \\
  
        OpenOccupancy~\cite{OpenOccupancy}  & C &19.3  & 10.3  &  9.9 & 6.8  & 11.2  & 11.5  & 6.3  & 8.4  & 8.6 & 4.3 & 4.2 & 9.9 & 22.0  & 15.8 & 14.1  & 13.5  & 7.3&10.2 &1716G & 19G & 0.84/1.22s \\
        
        C-CONet~\cite{OpenOccupancy} & C & 20.1  & 12.8&13.2  & 8.1 &  \textbf{15.4} &  17.2 & 6.3  & \textbf{11.2}  & 10.0  &  8.3 & 4.7 & 12.1 & 31.4 & 18.8 & 18.7  & 16.3 & 4.8  &8.2 &1810G & 21G & 2.18/2.58s\\
        
        \hline
        SparseOcc (ours) & C & \textbf{21.8} & \textbf{14.1} & \textbf{16.1} & \textbf{9.3} & 15.1 & \textbf{18.6} & \textbf{7.3} & 9.4 & \textbf{11.2} & \textbf{9.4} & \textbf{7.2} & \textbf{13.0} & \textbf{31.8} & \textbf{21.7} & \textbf{20.7} & \textbf{18.8} & {6.1} & \textbf{10.6} & 455G & 13G & 0.19/0.25s\\
        
	\hline
	\end{tabular}}\\
        \vspace{-8pt}
        \caption{\textbf{Semantic occpancy prediction results on nuScenes-Occupancy~\cite{OpenOccupancy} validation set.} For accuracy evaluation, We report the geometric metric IoU, semantic metric mIoU, and the IoU for each semantic class. For efficiency evaluation, we report the FLOPs, training GPU memory, and 3D/overall inference latency. The C denotes camera and the \textbf{bold} numbers indicate the best results.}
        \vspace{-5pt}
	\label{table:nusc}
\end{table*}
\begin{table*}
    \scriptsize
    \setlength{\tabcolsep}{0.0035\linewidth}
    \newcommand{\classfreq}[1]{{~\tiny(\semkitfreq{#1}\%)}}  %
    \centering
    \resizebox{1\linewidth}{!}{
    \begin{tabular}{l|c|c c| c c c c c c c c c c c c c c c c c c c|c}
        \hline
        Method & Input & IoU & mIoU
        & \rotatebox{90}{road\classfreq{road}} 
        & \rotatebox{90}{sidewalk\classfreq{sidewalk}}
        & \rotatebox{90}{parking\classfreq{parking}} 
        & \rotatebox{90}{other-ground\classfreq{otherground}} 
        & \rotatebox{90}{building\classfreq{building}} 
        & \rotatebox{90}{car\classfreq{car}} 
        & \rotatebox{90}{truck\classfreq{truck}} 
        & \rotatebox{90}{bicycle\classfreq{bicycle}} 
        & \rotatebox{90}{motorcycle\classfreq{motorcycle}} 
        & \rotatebox{90}{other-vehicle\classfreq{othervehicle}} 
        & \rotatebox{90}{vegetation\classfreq{vegetation}} 
        & \rotatebox{90}{trunk\classfreq{trunk}} 
        & \rotatebox{90}{terrain\classfreq{terrain}} 
        & \rotatebox{90}{person\classfreq{person}} 
        & \rotatebox{90}{bicyclist\classfreq{bicyclist}} 
        & \rotatebox{90}{motorcyclist\classfreq{motorcyclist}} 
        & \rotatebox{90}{fence\classfreq{fence}} 
        & \rotatebox{90}{pole\classfreq{pole}} 
        & \rotatebox{90}{traffic-sign\classfreq{trafficsign}} 
        & FLOPs\\
        \hline\hline
        LMSCNet*~\cite{roldao2020lmscnet} & C & 28.61 & 6.70 & 40.68 & 18.22 & 4.38 & 0.00 & 10.31 & 18.33 & 0.00 & 0.00 & 0.00 & 0.00 & 13.66 & 0.02 & 20.54 & 0.00 & 0.00 & 0.00 & 1.21 & 0.00 & 0.00 & -  \\ %
        3DSketch*~\cite{3d-sketch} & C & 33.30 & 7.50 & 41.32 & 21.63 & 0.00 & 0.00 & 14.81 & 18.59 & 0.00 & 0.00 & 0.00 & 0.00 & 19.09 & 0.00 & 26.40 & 0.00 & 0.00 & 0.00 & 0.73 & 0.00 & 0.00 & - \\ %
        AICNet*~\cite{li2020anisotropic} & C & 29.59 & 8.31 & 43.55 & 20.55 & 11.97 & 0.07 & 12.94 & 14.71 & 4.53 & 0.00 & 0.00 & 0.00 & 15.37 & 2.90 & 28.71 & 0.00 & 0.00 & 0.00 & 2.52 & 0.06 & 0.00 & -  \\ %
        JS3C-Net*~\cite{js3cnet} & C & \textbf{38.98} & 10.31 & 50.49 & 23.74 & 11.94 & 0.07 & 15.03 & 24.65 & 4.41 & 0.00 & 0.00 & 6.15 & 18.11 & \textbf{4.33} & 26.86 & 0.67 & 0.27 & 0.00 & 3.94 & 3.77 & 1.45  & -\\
        MonoScene$\dagger$~\cite{monoscene} & C & 36.86 & 11.08 & 56.52 & 26.72 & 14.27 & 0.46 & 14.09 & 23.26 & 6.98 & 0.61 & 0.45 & 1.48 & 17.89 & 2.81 & 29.64 & 1.86 & 1.20 & 0.00 & 5.84 & 4.14 & 2.25 & -\\
        TPVFormer~\cite{tpvformer} & C & 35.61 & 11.36 & 56.50 & 25.87 & \textbf{20.60} & \textbf{0.85} & 13.88 & 23.81 & 8.08 & 0.36 & 0.05 & 4.35 & 16.92 & 2.26 & 30.38 & 0.51 & 0.89 & 0.00 & 5.94 & 3.14 & 1.52 & 946G\\
        OccFormer~\cite{occformer} & C & 36.50 & \textbf{13.46} & 58.85 & 26.88 & 19.61 & 0.31 & 14.40 & \textbf{25.09} & \textbf{25.53} & \textbf{0.81} & \textbf{1.19} & 8.52 & \textbf{19.63} & 3.93 & \textbf{32.62} & 2.78 & \textbf{2.82} & 0.00 & 5.61 & \textbf{4.26} & \textbf{2.86} & 889G \\
        \midrule
        SparseOcc & C & 36.48 & 13.12 & \textbf{59.59} & \textbf{29.68} & 20.44 & 0.47 & \textbf{15.41} & 24.03 & 18.07 & 0.78 & 0.89 & \textbf{8.94} & 18.89 & 3.46 & 31.06 & \textbf{3.68} & 0.62 & 0.00 & \textbf{6.73} & 3.89 & 2.60 & 393G \\
        \hline
    \end{tabular}}\\
    \vspace{-8pt}
    \caption{\textbf{Semantic scene completion results on SemanticKITTI~\cite{behley2019semantickitti} validation set.} For accuracy evaluation, We report the geometric metric IoU, semantic metric mIoU, and the IoU for each semantic class. For efficiency evaluation, we report the FLOPs. The C denotes camera and the \textbf{bold} numbers indicate the best results. The methods with ``*" are RGB-input variants reported by~\cite{monoscene} for fair comparison. }
    \label{table:kitti}
    \vspace{-10pt}
\end{table*}
\subsection{Experimental Setup}
\noindent\textbf{Datasets.}
We evaluate our proposed SparseOcc on {nuScenes-Occupancy}~\cite{OpenOccupancy} and {SemanticKITTI}~\cite{semantickitti}. 
\textbf{nuScenes-Occupancy}~\cite{OpenOccupancy} extends the famous large-scale autonomous driving datasets {nuScenes}~\cite{nuscenes} with 3D dense semantic occupancy annotations on key frames for 1 ``empty" class and 16 semantic classes using Augmenting And Purifying pipeline. It covers 700 and 150 driving scenes in the training and validation set of nuScenes. The multi-view images and corresponding LiDAR points are provided by the original nuScenes dataset. 
\textbf{SemanticKITTI}~\cite{semantickitti} contains 22 sequences including monocular images, LiDAR points, point cloud segmentation labels and semantic scene completion annotations. The sequence 08 is officially split for validation, sequences 00-10 (excluding 08) are used for training, and sequences 11-21 are the test set. It annotates each 3D voxel with 1 ``empty" class or 19 semantic classes.

\noindent\textbf{Evaluation Metric.}
We follow the common practice and report the intersection over union (IoU) of occupied voxels regardless of their semantic labels to evaluate the reconstructed geometry shape. The semantic mIoU of all semantic classes is also reported to evaluate the semantic-aware perception ability. For efficiency analysis, FLOPs is also evaluated using the analysis tool provided by~\cite{bevfusion}.

\noindent\textbf{Implementation Details.}
For image encoder, we follow the previous work~\cite{occformer, OpenOccupancy} for fair comparison and use EfficientNetB7~\cite{efficientnet} and SecondFPN~\cite{second} on SemanticKITTI and ResNet-50~\cite{resnet} on nuScenes-Occupancy. The 2D to 3D view transformation is implemented as the same as~\cite{occformer,OpenOccupancy}. And it generates a 3D feature volume of size 128$\times$128$\times$16 and 128$\times$128$\times$10 with 128 channels for SemanticKITTI and nuScenes-Occupancy, which is converted to sparse representation. We stack the sparse diffuser for $L=4$ times, each followed by a sparse convolution with stride 2 for downsampling. The kernel size $k$ of the orthogonal convolutions in sparse completion block and contextual aggregation block is set to 3. The sparse voxel decoder projects the multi-scale features to 192 channels and enhances them via interpolation and summation. The sparse transformer head consists of 9 layers for query updating and decoding and the number of queries $N_q$ is 100. Following~\cite{occformer,mask2former}, we sample 50176 points for supervision according to the class frequency for fast training speed.
During inference, the predicted masks are upsampled 2$\times$ and 4$\times$ to the size of ground-truth via trilinear interpolation for evaluation on SemanticKITTI and nuScenes-Occupancy, respectively. 

\noindent\textbf{Optimization.}
AdamW~\cite{adamw} optimizer with an initial learning rate of 1e-4 and a weight decay of 0.01 is leveraged during training. We train the model for 30 epochs on SemanticKITTI and 24 epochs on nuScenes-Occupancy with a batch size of 8 on 8 v100 GPUs.

\subsection{Benchmark Performance}
In this part, we compare our SparseOcc with the published state-of-the-art methods on nuScenes-Occupancy and SemanticKITTI. The results of the former methods are directly derived from their papers or code with their configurations.

\noindent\textbf{NuScenes-Occupancy.}
As shown in Tab.~\ref{table:nusc}, our SparseOcc successfully outperforms the 3D dense representation based C-CONet~\cite{OpenOccupancy} by 1.7\% geometry IoU and 1.3\% semantic mIoU. This accomplishment highlights SparseOcc's effectiveness. Besides, we can observe that the model FLOPs are reduced by 74.9\% relatively, which further justifies the efficiency of our SparseOcc. 
More surprisingly, SparseOcc not only produces a remarkable 80.8\% improvement in mIoU over TPVFormer~\cite{tpvformer} which uses efficient TPV view as 3D scene representation, but also reduces the FLOPs by 59.8\%. 
We find that TPVFormer uses a relatively large image backbone ResNet-101~\cite{resnet}, and the multi-layer image cross-attention as well as cross-view attention between dense TPV maps also impose lots of computational burden. 
In contrast, our SparseOcc uses sparse operators in the 3D space, helping us achieve superior performance with a lightweight implementation.
Besides, SparseOcc exhibits a noteworthy reduction in 3D inference latency, i.e., the time for 3D feature processing.

\noindent\textbf{SemanticKITTI.}
We quantitatively compare the proposed SparseOcc with several previous works in Tab.~\ref{table:kitti}. we can see that SparseOcc achieves comparable performance with OccFormer~\cite{occformer} and significantly better than other methods in terms of semantic mIoU. Please note that different from OccFormer which builds long-range and dynamic dependency in the 3D space using Swin Transformer~\cite{swin_transformer}, the 3D encoder of SparseOcc is built with spconv~\cite{spconv} only. Consequently, the model FLOPs of SparseOcc is just 44.2\% of OccFormer~\cite{occformer} approximately.  

\subsection{Qualitative Evaluation}
\begin{figure*}
    \centering
    \includegraphics[scale=.53]{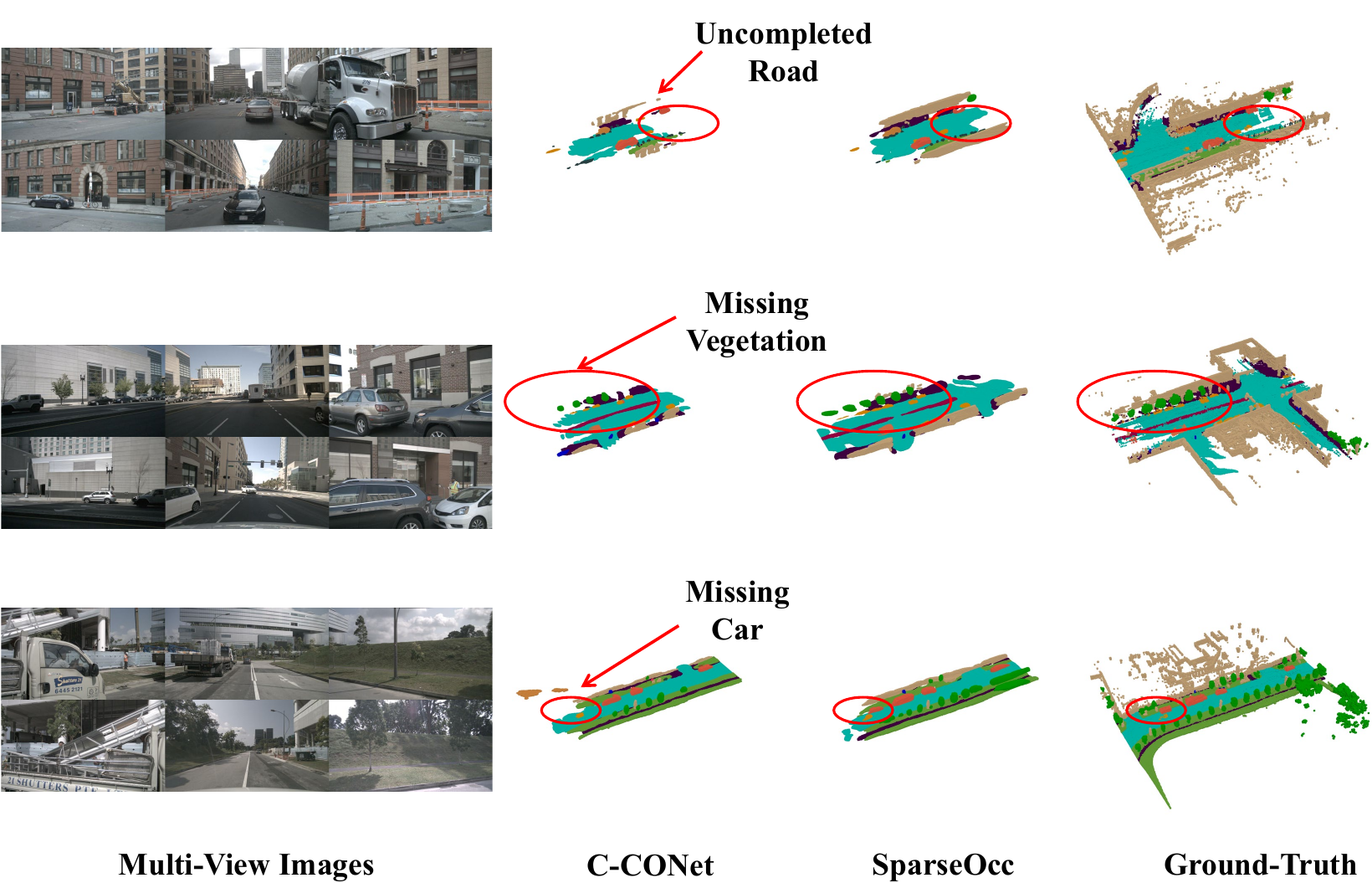}
    \vspace{-7pt}
    \caption{\textbf{Qualitative results of 3D semantic occupancy on nuScenes-Occupancy validation set.} The input multi-view images are shown on the leftmost and the occupancy predictions of C-CONet~\cite{OpenOccupancy}, our SparseOcc, and the ground-truth are then visualized sequentially. Compared to 3D dense representation based C-CONet~\cite{yan20222dpass}, our SparseOcc achieves better completion and segmentation as highlighted by the red circles.}
    \label{fig:res_vis}
    \vspace{-10pt}
\end{figure*}
We visualize the 3D semantic occupancy prediction results of several challenging scenes in nuScenes-Occupancy validation set in Fig.~\ref{fig:res_vis}. As can be seen, compared with the 3D dense counterpart C-CONet~\cite{OpenOccupancy}, our SparseOcc can better complete the large area flat-like road (first row), capture the complicated structure of vegetation (second row) and reconstruct car in the distance (third row). We can conclude that benefiting from the proposed sparse latent diffuser and learned sparse feature pyramid, our sparse occupancy transformer head can generate relatively accurate scene-level descriptions in an efficient way.

\subsection{Ablation Studies}
\begin{table}[t]
  \centering
  \resizebox{0.42\textwidth}{!}{
    \begin{tabular}{c|c|c|c|c}
         \hline
          Conv Type &  Conv Num & Kernel Size &  IoU & mIoU\\
        \hline\hline
       \noalign{\smallskip}
       - & 0 & - & 35.5 & 12.1 \\
       \hline
       \multirow{3}{*}{Regular}
           & 1 & \multirow{3}{*}{3$\times$3$\times$3} &  35.8 & 12.2 \\
           \cline{2-2}\cline{4-5}
           & 2  & & 36.4 & 12.3\\
           \cline{2-2}\cline{4-5}
           & 3 & & 36.2 & 12.6\\
        \hline
        \multirow{3}{*}{Decomposed}
           &  1  & 3$\times$3$\times$1 & 36.5 & \textbf{13.1}\\ 
           \cline{2-2}\cline{4-5}
           &  2  & 3$\times$1$\times$3 & \textbf{36.6} & 12.8 \\
           \cline{2-2}\cline{4-5}
           &  3  & 1$\times$3$\times1$ & 36.4 & 12.7 \\
         \hline
    \end{tabular}}\
    \vspace{-7pt}
  \caption{Ablation on different designs of Sparse Completion Block on SemanticKITTI val set.} 
    \label{tab:abl_scb} 
    \vspace{-7pt}
\end{table}
\noindent\textbf{Sparse Completion Blcok.}
As illustrated in Sec.~\ref{sec:sparse_latent_diffuser}, a 3D diffusion kernel is spatially decomposed into a combination of three orthogonal convolutional kernels in the sparse completion block. Tab.~\ref{tab:abl_scb} ablates the type and number of diffusion kernels. Surprisingly, we find that SparseOcc can perform satisfactorily even if no sparse completion block is applied. We think that it is because the decoder which fuses sparse feature pyramid can also complete the scene to some extent. Compared with regular 3$\times$3$\times$3 kernel, the decomposed orthogonal achieves better scene completion and segmentation results. 
Besides, stacking more convolution blocks does not improve performance. Hence, we build our sparse completion block with only a group of decomposed orthogonal. 

\noindent\textbf{Sparse Feature Pyramid.}
\begin{table}[t]
  \centering
  \resizebox{0.42\textwidth}{!}{
    \begin{tabular}{c|c|c|c|c}
         \hline
          Type  &  IoU & mIoU & Memory & FLOPs\\
        \hline\hline
       FPN3D\cite{fpn} & 34.4 & 9.8 & 13.2G & 307G\\
       \hline
       MSDeformAttn3D & \textbf{36.7} & \textbf{13.3} & 19.8G & 379G\\
       \hline
       Sparse Decoder & 36.5 & 13.1 & 13.3G & 279G\\ 
       \hline
    \end{tabular}}
    \vspace{-7pt}
  \caption{Ablation on voxel decoder on SemanticKITTI val set.} 
    \label{tab:abl_vox_dec} 
    \vspace{-7pt}
\end{table}
As shown in Tab.~\ref{tab:abl_vox_dec}, we observe that when the SparseOcc equipped with 6 layers of multi-scale deformable attention (MSDeformAttn), it performs better than the proposed sparse decoder. However, when the layer number of MSDeformAttn decreases, the IoU and mIoU both drop and lag behind the proposed simple sparse decoder. Moreover, the training GPU memory and FLOPs of the MSDeformAttn3D are much higher than the proposed interpolation and summation method. Compared with FPN3D~\cite{fpn}, our sparse voxel decoder is 2.1\% IoU and 3.3\% mIoU higher with slightly more memory. 

\noindent\textbf{Sparse Transformer Head.}
\begin{table}[t]
  \centering
  \resizebox{0.42\textwidth}{!}{
    \begin{tabular}{c|c|c|c|c}
         \hline
          Type &   IoU & mIoU & Memory & FLOPs\\
        \hline\hline
        Linear Head & \textbf{36.8} & 11.8 & 9.8G & 5.4G\\
        Trans. Head & 36.2 &  \textbf{13.2} & 19.9G & 19.0G\\
       \hline
        Sparse Trans. Head  & 36.5 & 13.1 & 13.3G & 13.5G \\ 
           
         \hline
    \end{tabular}}
    \vspace{-7pt}
  \caption{Ablation on segmentation head on SemanticKITTI val set.} 
    \label{tab:abl_trans_dec} 
    \vspace{-7pt}
\end{table}
We compare several different prediction heads, including simple linear head, transformer head proposed in~\cite{occformer}, and our sparse transformer head in Tab.~\ref{tab:abl_trans_dec}. As can be seen, the linear head achieves the best geometry IoU of 36.8. We postulate it may be because the output of the linear head is supervised by an explicit geometry loss, i.e., a cross-entropy loss on occupied and non-occupied voxels. On the contrary, the transformer decoder formulates occupancy prediction as mask generation for each semantic class exclusively. Consequently, no explicit supervision is or can be imposed on the occupied voxels. SparseOcc uses a linear layer for coarse segmentation to filter out the non-occupied voxels and then proposes a sparse transformer head for mask prediction. Therefore, it makes an appropriate trade-off between these two kinds of occupancy heads and achieves satisfactory performance with low training memory.

\noindent\textbf{Input Image Size.}
\begin{table}[t]
  \centering
  \resizebox{0.45\textwidth}{!}{
    \begin{tabular}{c|c|c|c|c}
         \hline
          Method &  2D Backbone &  Input Size & IoU & mIoU\\
        \hline\hline
       \multirow{3}{*}{
       C-CONet~\cite{OpenOccupancy}}
           & R-50 & $704\times 256$ &16.6 &8.6 \\
           & R-50 & $1600\times 900$ & 19.3 & 10.3\\
           & R-101 & $1600\times 900$ & 20.2 & 11.4\\
        \hline
        \multirow{2}{*}{SparseOcc}
          &  R-50  & $704\times 256$ &  \textbf{21.8} & 14.1\\ 
          
           &  R-50  & $1600\times 900$ & 20.4 & \textbf{14.6} \\           
         \noalign{\smallskip}
         \hline
    \end{tabular}}
    \vspace{-7pt}
  \caption{Ablation on input size and 2D backbone on nuScenes-Occupancy.} 
    \label{tab:abl_img_size} 
    \vspace{-10pt}
\end{table}
A larger input image size has a larger resolution of image features, thus having a larger feature density in the 3D space, which further influences the performance. From Tab.~\ref{tab:abl_img_size}, we can observe that SparseOcc still performs better than C-CONet, even though it is trained using a smaller input size (704$\times$256) and image backbone (R-50), which further demonstrates the effectiveness of our SparseOcc. Additionally, using a larger input image size (1600$\times$900) can improve mIoU for both C-CONet and the proposed SparseOcc, thanks to the density increase of semantic features. 
However, the IoU of SparseOcc drops when using higher image resolution. We blame it on the wrong hallucination on spatially empty voxels caused by over-dense 3D sparse features. When inactivating the sparse completion block in the last sparse latent diffuser layer, this hallucination can be relieved. 
\section{Conclusion}
In this paper, we explore the possibility of using pure sparse representation for 3D scene description and present SpaseOcc for 3D semantic occupancy prediction. Specifically, a sparse latent diffuser with decomposed orthogonal kernels is proposed to propagate the non-empty features to their adjacent empty area. To efficiently complete the scene, we stack layers of diffusers and downsampling layers to generate a coarse-to-fine sparse feature pyramid, which is further fused via simple interpolation and summation. In this way, the feature pyramid is enhanced by multi-scale semantic information. Finally, we use a sparse transformer head to query multi-scale sparse features and generate semantic occupancy predictions. Without bells and whistles, SparseOcc achieves state-of-the-art results on nuScenes-Occupancy and comparable performance on SemanticKITTI, fully demonstrating its superior effectiveness and efficiency. 

\smallskip
\noindent\textbf{Acknowledgements.} This work was supported by NSFC (62322113, 62376156), Shanghai Municipal Science and Technology Major Project (2021SHZDZX0102), and the Fundamental Research Funds for the Central Universities.

{
    \small
    \bibliographystyle{ieeenat_fullname}
    \bibliography{main}
}


\end{document}